\documentclass[twocolumn]{bytedance_seed}

\usepackage{amsmath}
\usepackage{amssymb}
\usepackage{tikz}
\usetikzlibrary{positioning,arrows.meta,calc}
\usepackage{array}
\usepackage{url}
\usepackage{enumitem}

\setlist[itemize]{leftmargin=*,topsep=2pt,itemsep=1pt,parsep=0pt}
\microtypesetup{expansion=false}
\renewcommand{\arraystretch}{1.08}

\newcommand{\hcmult}{m}
\newcommand{\hc}{HC}
\newcommand{\hcfull}{Hyper-Connections (\hc{})}
\newcommand{\prehc}{\texttt{pre\_hc\_head}}
\newcommand{\hchead}{\texttt{hc\_head}}
\newcommand{\fc}{\texttt{fc}}

\newcommand{\best}[1]{{\bfseries\boldmath #1}}

\title{HyperDFlash: Hyper-Connection-Aligned Block Speculative Decoding with Gated Residual Reduction}

\author[1]{Luxi Lin$^{*}$, Shuang Peng$^{*,\dagger}$, Rui Ma, Junhao Hua, Shuwei Fan, Zhengda Qin, Qiang Wang, Hongjian Sun, Fangmin Chen$^{\ddagger}$, Songwei Liu$^{\ddagger}$}
\affiliation[1]{ByteDance}
\correspondence{\email{liusongwei.zju@bytedance.com}}

\abstract{

We present HyperDFlash, a block-parallel speculative decoding framework tailored to DeepSeek-V4's \hcfull{}~\cite{xu2026deepseek,zhu2024hyper}.
Despite the strong performance of DeepSeek-V4’s native Multi-Token Prediction (MTP) module on initial token drafting, its draft accuracy degrades sharply at later positions, as error accumulation from unverified intermediate tokens severely harms draft acceptance rates.
Although the original DFlash method supports efficient one-pass block drafting, it cannot be seamlessly adapted to the \hc{} paradigm, since DeepSeek-V4's $\hcmult$-path residual stream induces inherent feature misalignment with conventional drafting designs.
To resolve this architectural mismatch, we propose two dedicated, model-aligned optimizations for \hc{} residual streams.
First, we adopt pre-collapse residual states as the exclusive conditioning signal, preserving complete multi-path structural information and better aligning the drafter with the target’s native prediction pathway.
Second, we replace the heavy generic linear compressor with a lightweight gated residual reducer, whose parameters are directly inherited from the target model’s built-in \hchead{} module. This design yields input-aware path aggregation with three orders of magnitude fewer parameters while maintaining precise architectural alignment.
We further enhance model training via a targeted KL distillation loss applied to the LM-head. This objective regularizes model predictions against the full probability distribution of the target model, thereby boosting the quality of draft sequences at early training stages.
Extensive experiments across math reasoning, code synthesis, and conversational benchmarks demonstrate that HyperDFlash consistently outperforms both the native MTP baseline and vanilla DFlash adaptation. 
It achieves substantial gains in average accepted draft length and decoding speedup, validating the effectiveness of \hc{} alignment, efficient gated reduction, and targeted distillation for high-performance speculative decoding.
}

\begin{document}
\maketitle
\begingroup
\renewcommand{\thefootnote}{}
\footnotetext{$^{*}$Equal contribution. $^{\dagger}$Project lead. $^{\ddagger}$Corresponding author.}
\endgroup

\begin{figure*}[t]
\centering
\includegraphics[width=0.45\textwidth]{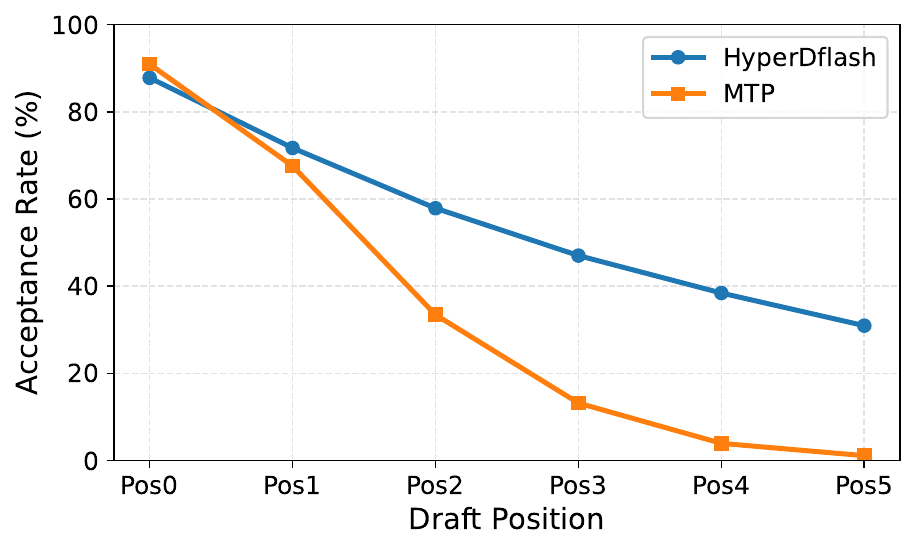}
\caption{Per-position acceptance rates over six drafted positions.
Native MTP achieves high acceptance at the first position but degrades rapidly at later positions, while HyperDFlash maintains a smoother acceptance profile.}
\label{fig:position-acceptance}
\end{figure*}

\section{Introduction}
\label{sec:intro}

Speculative decoding accelerates generation by letting a lightweight draft model propose future tokens that the target model verifies in parallel~\cite{leviathan2023,chen2023}. 
The attainable speedup is determined not only by whether the first drafted token is correct, but also by how many consecutive draft tokens can be accepted. 
This makes long-range draft accuracy especially important.

DeepSeek-V4 includes a native Multi-Token Prediction (MTP) module~\cite{gloeckle2024,xu2026deepseek}, which provides a natural drafter for speculative decoding. 
However, due to its sequential training and inference formulation, MTP maintains high acceptance only at the first one or two draft positions; the acceptance rate drops sharply at later positions as the module extrapolates from unverified draft tokens. 
As illustrated in Figure~\ref{fig:position-acceptance}, this positional degradation limits the accepted prefix length and therefore the practical acceleration potential of native MTP.

To address this limitation, we adopt DFlash~\cite{chen2026dflash} as a draft model. 
Unlike autoregressive MTP-style drafting, DFlash predicts an entire draft block from an accepted anchor token and target hidden-state context, avoiding the accumulation of errors along unverified draft tokens. 
However, directly transferring DFlash to DeepSeek-V4 is non-trivial because DeepSeek-V4 adopts \hcfull{}~\cite{xu2026deepseek,zhu2024hyper}: each token is represented by $\hcmult$ parallel residual paths before prediction, rather than by a single hidden state.
These paths are collapsed by the target model's own \hchead{} module immediately before the LM head. Therefore, a generic DFlash adaptation faces two sources of mismatch: using intermediate-layer features moves the conditioning signal away from the final prediction state, while reducing the multi-path residual with a dense learned projection ignores the target model's native path-aggregation mechanism.

%

HyperDFlash instead reuses the final pre-collapse residual already stored for MTP, aligns the drafting signal with the target \hc{} prediction pathway via a lightweight inherited gated reducer, and further enhances draft token quality with a targeted KL distillation objective. Focused experiments validate that our \hc{}-consistent designs substantially improve accepted token length and generation speed over vanilla DFlash and native MTP across math, code, and conversational benchmarks.

\begin{figure*}[t]
\centering
\includegraphics[width=0.68\textwidth]{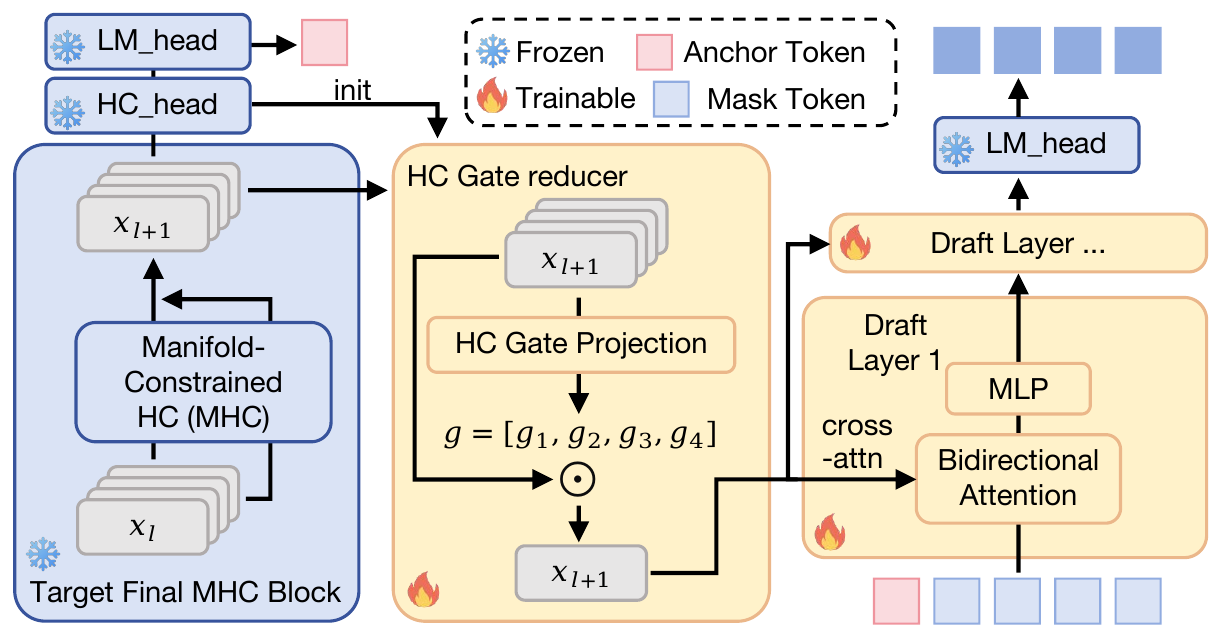}
\caption{Overview of HyperDFlash.
The drafter conditions on the target model's final pre-collapse \hc{} residual, which is reduced by the proposed Inherited HC-Gate Reducer and injected into a lightweight DFlash drafter.
An auxiliary KL distillation objective further improves early-position draft quality.}
\label{fig:overview}
\end{figure*}

\section{Method}
\label{sec:method}

Figure~\ref{fig:overview} illustrates the overall pipeline of HyperDFlash. We first establish consistent conditioning by leveraging the target model's pre-collapse \hc{} residual states (Section~\ref{sec:source}). Then, we introduce an inherited gated reducer to achieve parameter-efficient and model-aligned path collapsing (Section~\ref{sec:reduce}). Finally, we adopt a targeted LM-head distillation strategy to refine early-position draft predictions during training (Section~\ref{sec:kl}).

\subsection{Pre-Collapse Residual Conditioning for Hyper-Connection Consistency}
\label{sec:source}
A core challenge in adapting speculative drafting to DeepSeek-V4’s \hc{} architecture lies in bridging the contextual gap between the drafter and the target prediction pipeline. Conventional drafters rely on aggregated intermediate layer features, which incur redundant computation and break structural alignment with the \hc{} multi-path residual stream. Instead, we use the target’s final pre-collapse \hc{} residual states---the \prehc{} source---as the target-side conditioning feature for block drafting, while the drafter still follows the standard anchor-plus-mask block input format.
This preserves the native multi-path semantics and improves consistency with the target’s prediction pathway.

\subsection{Inherited HC-Gate Reducer for Efficient Path Collapsing}
\label{sec:reduce}
DeepSeek-V4’s \hc{} design maintains multiple parallel residual paths per token, which cannot be properly aggregated by conventional static linear projections. A naive dense collapse introduces severe structural mismatch and enormous parameter overhead, breaking consistency between drafting and target inference.
To tackle this issue, we design an Inherited HC-Gate Reducer that follows the functional form of the target’s native \hchead{} aggregation mechanism. By initializing the gate parameters from the pretrained target \hchead{} weights, our method achieves structurally aligned residual collapsing with three orders of magnitude fewer parameters.

\begin{table*}[t]
\caption{Design comparison of the path-collapse reducers for DeepSeek-V4-Flash.
The Inherited HC-Gate Reducer follows the input-dependent \hc{} path-gating form of the target model, delivering high parameter efficiency via architectural alignment rather than standalone feature compression.}
\label{tab:reducer}
\centering
\scriptsize
\begin{tabular}{@{}ccc@{}}
\toprule
 & Generic \fc{} Reducer & Inherited HC-Gate Reducer \\
\midrule
Collapse Mechanism & Dense Linear Projection & Gated Path-wise Aggregation \\
Trainable Parameters & 67M & 65K \\
Weight Initialization & Random Initialization & Inherited from Target \hchead{} \\
Target Alignment & Learned from Scratch & Inherently Aligned with Target Model \\
\bottomrule
\end{tabular}
\end{table*}

A generic DFlash reducer uses a static $\fc:\mathbb{R}^{\hcmult d}\!\rightarrow\!\mathbb{R}^{d}$ projection ($\sim$67M parameters for DeepSeek-V4-Flash).
This treats the flattened \hc{} residual as an ordinary long vector and learns a dense reduction from scratch.
However, the target model itself does not collapse its residual paths with a static linear map.
Instead, the target's own \hchead{} collapse is \emph{input-dependent}: for token $t$, it first normalizes the flattened multi-path residual and then computes one scalar gate per path,

\begin{equation}
\begin{aligned}
\tilde{\mathbf{x}}_t &= \mathrm{RMSNorm}\!\big(\mathrm{vec}(\mathbf{H}_t)\big),\\
\boldsymbol{\alpha}_t &= \sigma\!\big(W_f\,\tilde{\mathbf{x}}_t + b\big),\\
\mathbf{y}_t &= \sum_{j=1}^{\hcmult} \alpha_{t,j}\,\mathbf{H}_{t,j},
\end{aligned}
\label{eq:hchead}
\end{equation}
where $W_f\in\mathbb{R}^{\hcmult\times \hcmult d}$ and each $\alpha_{t,j}$ is a scalar sigmoid gate applied to the entire residual path $\mathbf{H}_{t,j}$.
The gates are independent per-path sigmoids rather than a softmax distribution over paths.
In implementation, the target module uses its original inverse-RMS scaling, scale factor, and small numerical offset; Eq.~\ref{eq:hchead} abstracts this implementation into its input-dependent path-gating form.
A static linear map cannot express this input-dependent gating, so it is misaligned with the target from the start.

We instantiate a drafter-side reducer using the same input-dependent path-gating form as Eq.~\ref{eq:hchead}, with an additional lightweight per-path RMSNorm for the drafter input, and initialize its gate parameters from the target's \hchead{} weights.
The draft model thus starts from a collapse already aligned with the target's prediction pathway; the large parameter reduction is a by-product of this alignment, not the goal (Table~\ref{tab:reducer}).

The reducer is specialized: it assumes the \prehc{} source and matching target and draft hidden widths.
When either assumption fails, the generic \fc{} projection remains the fallback.

\subsection{LM-Head KL Distillation for Early-Stage Draft Refinement}
\label{sec:kl}
Standard cross-entropy (CE) loss performs one-hot token supervision based on discrete label signals and fails to exploit the full probabilistic distribution modeled by the target language model. To enhance the regularization of block-parallel drafting, we incorporate an auxiliary KL distillation objective. This scheme produces soft supervisory signals from the target’s LM head on cached hidden states without additional forward propagation overhead. In block-wise drafting, later prediction steps suffer from inconsistent contextual information due to masked intermediate tokens. Accordingly, we restrict distillation supervision to early block steps. This targeted regularization stabilizes model training and effectively improves the prediction quality of critical early draft tokens.

Since the target hidden state $\mathbf{h}_p$ at position $p$ encodes the context $[0{:}p]$, $\mathrm{LMHead}(\mathbf{h}_p)$ predicts the token at $p{+}1$.
For a block anchored at position $a$, draft position $k$ (which predicts the token at $a{+}k$) is therefore aligned as
\begin{equation*}
\text{teacher}_k = \mathrm{LMHead}\big(\mathbf{h}_{a+k-1}\big),\quad
\text{student}_k = \mathbf{z}_k,
\end{equation*}
where $\mathbf{z}_k$ denotes the draft logits.
With distillation temperature $T_{\mathrm{KD}}$ and the first $P$ block positions supervised,
\begin{align}
\mathcal{L}_{\mathrm{KL}} &= \frac{T_{\mathrm{KD}}^2}{P}\sum_{k=1}^{P}
\mathbb{E}\Big[\mathrm{KL}\big(p^{T_{\mathrm{KD}}}_k \,\|\, q^{T_{\mathrm{KD}}}_k\big)\Big],\\
\mathcal{L} &= \mathcal{L}_{\mathrm{CE}} + \alpha\,\mathcal{L}_{\mathrm{KL}},
\label{eq:loss}
\end{align}
where $p^{T_{\mathrm{KD}}}_k$ and $q^{T_{\mathrm{KD}}}_k$ are the softmax distributions of the teacher and draft logits scaled by $1/T_{\mathrm{KD}}$.
The $T_{\mathrm{KD}}^2$ factor follows standard distillation practice~\cite{hinton2015}.

This teacher is most reliable at $k{=}1$: both teacher and student condition on the same context $[0{:}a]$, so the teacher simply provides the full target distribution that the CE one-hot label omits.
For $k{\geq}2$, the teacher also conditions on the ground-truth intermediate tokens $a{+}1,\dots,a{+}k{-}1$, whereas the drafter sees only mask tokens at those slots and should instead model the marginal distribution over them; matching the teacher's sharp, per-example conditionals then yields high-variance, conflicting gradients.
We therefore apply KL only to the first two positions with a conservative $\alpha$ (typically $0.1$--$0.2$).
In the current implementation, the KL teacher mean-pools the \hc{} paths rather than using the gated \hchead{} collapse, which gives us a further reason to keep the weight of this auxiliary term small.

\begin{table*}[t]
\caption{Main decoding results for DeepSeek-V4-Flash in \best{Non-thinking} mode.
We report decoding speedup over target-only autoregressive decoding and average accepted length ($\tau$).
Parenthesized values indicate the number of drafted steps per verification round.}
\label{tab:main-results-nonthink}
\centering
\resizebox{\linewidth}{!}{
\scriptsize
\setlength{\tabcolsep}{2pt}
\renewcommand{\arraystretch}{0.92}
\begin{tabular}{p{2.7cm} @{\hspace{1.2em}} cc cc cc @{\hspace{1.2em}} cc cc cc @{\hspace{1.2em}} cc @{\hspace{1.2em}} cc}
\toprule
\multirow{2}{*}{Method} &
\multicolumn{6}{c@{\hspace{1.2em}}}{\sc{Math}} &
\multicolumn{6}{c@{\hspace{1.2em}}}{\sc{Code}} &
\multicolumn{2}{c@{\hspace{1.2em}}}{\sc{Chat}} &
\multicolumn{2}{c}{\textit{Avg.}} \\
\cmidrule(lr){2-7}\cmidrule(lr){8-13}\cmidrule(lr){14-15}\cmidrule(l){16-17}
& \multicolumn{2}{c}{GSM8K}
& \multicolumn{2}{c}{MATH-500}
& \multicolumn{2}{c@{\hspace{1.2em}}}{AIME25}
& \multicolumn{2}{c}{HumanEval}
& \multicolumn{2}{c}{MBPP}
& \multicolumn{2}{c@{\hspace{1.2em}}}{LCB}
& \multicolumn{2}{c@{\hspace{1.2em}}}{MT-Bench}
& \multicolumn{2}{c}{Average} \\
\midrule
Temperature = 0
& Speedup & $\tau$
& Speedup & $\tau$
& Speedup & $\tau$
& Speedup & $\tau$
& Speedup & $\tau$
& Speedup & $\tau$
& Speedup & $\tau$
& Speedup & $\tau$ \\
\midrule
MTP (3)
& 2.76$\times$ & 3.06
& 2.36$\times$ & 3.13
& 2.28$\times$ & 2.95
& 2.10$\times$ & 3.06
& 2.24$\times$ & 2.92
& 2.06$\times$ & 2.87
& 1.91$\times$ & 2.49
& 2.25$\times$ & 2.93 \\
MTP (6)
& 2.38$\times$ & 3.25
& 1.88$\times$ & 3.39
& 1.72$\times$ & 3.14
& 1.63$\times$ & 3.20
& 1.71$\times$ & 3.05
& 1.60$\times$ & 3.01
& 1.39$\times$ & 2.53
& 1.76$\times$ & 3.08 \\
Vanilla DFlash (6)
& 2.41$\times$ & 2.41
& 1.82$\times$ & 2.35
& 1.65$\times$ & 2.11
& 1.69$\times$ & 2.24
& 1.70$\times$ & 2.16
& 1.48$\times$ & 1.93
& 1.39$\times$ & 1.76
& 1.73$\times$ & 2.14 \\
HyperDFlash (6)
& \best{3.80$\times$} & \best{4.04}
& \best{2.93$\times$} & \best{3.95}
& \best{2.62$\times$} & \best{3.43}
& \best{2.74$\times$} & \best{4.29}
& \best{3.02$\times$} & \best{4.01}
& \best{2.38$\times$} & \best{3.40}
& \best{2.07$\times$} & \best{2.72}
& \best{2.80$\times$} & \best{3.69} \\
\midrule
Temperature = 1
& Speedup & $\tau$
& Speedup & $\tau$
& Speedup & $\tau$
& Speedup & $\tau$
& Speedup & $\tau$
& Speedup & $\tau$
& Speedup & $\tau$
& Speedup & $\tau$ \\
\midrule
MTP (3)
& 2.01$\times$ & 2.93
& 2.24$\times$ & 2.96
& 2.03$\times$ & 2.62
& 2.06$\times$ & 2.95
& 2.18$\times$ & 2.84
& 2.02$\times$ & 2.73
& 1.78$\times$ & 2.31
& 2.05$\times$ & 2.76 \\
MTP (6)
& 1.61$\times$ & 3.09
& 1.73$\times$ & 3.15
& 1.53$\times$ & 2.83
& 1.58$\times$ & 3.08
& 1.61$\times$ & 2.96
& 1.50$\times$ & 2.79
& 1.26$\times$ & 2.34
& 1.54$\times$ & 2.89 \\
Vanilla DFlash (6)
& 1.80$\times$ & 2.39
& 1.75$\times$ & 2.25
& 1.51$\times$ & 1.93
& 1.66$\times$ & 2.21
& 1.66$\times$ & 2.12
& 1.42$\times$ & 1.87
& 1.33$\times$ & 1.70
& 1.59$\times$ & 2.07 \\
HyperDFlash (6)
& \best{2.59$\times$} & \best{3.81}
& \best{2.72$\times$} & \best{3.67}
& \best{2.27$\times$} & \best{2.97}
& \best{2.67$\times$} & \best{4.17}
& \best{2.90$\times$} & \best{3.85}
& \best{2.17$\times$} & \best{3.16}
& \best{1.91$\times$} & \best{2.51}
& \best{2.46$\times$} & \best{3.45} \\
\bottomrule
\end{tabular}
}
\end{table*}

\begin{table*}[t]
\caption{Main decoding results for DeepSeek-V4-Flash in \best{Think-high} mode.
We report decoding speedup over target-only autoregressive decoding and average accepted length ($\tau$).
Parenthesized values indicate the number of drafted steps per verification round.}
\label{tab:main-results-thinkhigh}
\centering
\resizebox{\linewidth}{!}{
\scriptsize
\setlength{\tabcolsep}{2pt}
\renewcommand{\arraystretch}{0.92}
\begin{tabular}{p{2.7cm} @{\hspace{1.2em}} cc cc cc @{\hspace{1.2em}} cc cc cc @{\hspace{1.2em}} cc @{\hspace{1.2em}} cc}
\toprule
\multirow{2}{*}{Method} &
\multicolumn{6}{c@{\hspace{1.2em}}}{\sc{Math}} &
\multicolumn{6}{c@{\hspace{1.2em}}}{\sc{Code}} &
\multicolumn{2}{c@{\hspace{1.2em}}}{\sc{Chat}} &
\multicolumn{2}{c}{\textit{Avg.}} \\
\cmidrule(lr){2-7}\cmidrule(lr){8-13}\cmidrule(lr){14-15}\cmidrule(l){16-17}
& \multicolumn{2}{c}{GSM8K}
& \multicolumn{2}{c}{MATH-500}
& \multicolumn{2}{c@{\hspace{1.2em}}}{AIME25}
& \multicolumn{2}{c}{HumanEval}
& \multicolumn{2}{c}{MBPP}
& \multicolumn{2}{c@{\hspace{1.2em}}}{LCB}
& \multicolumn{2}{c@{\hspace{1.2em}}}{MT-Bench}
& \multicolumn{2}{c}{Average} \\
\midrule
Temperature = 0
& Speedup & $\tau$
& Speedup & $\tau$
& Speedup & $\tau$
& Speedup & $\tau$
& Speedup & $\tau$
& Speedup & $\tau$
& Speedup & $\tau$
& Speedup & $\tau$ \\
\midrule
MTP (3)
& 2.19$\times$ & 2.93
& 2.18$\times$ & 2.92
& 2.18$\times$ & 2.81
& 2.10$\times$ & 2.74
& 2.09$\times$ & 2.71
& 2.09$\times$ & 2.70
& 1.94$\times$ & 2.51
& 2.11$\times$ & 2.76 \\
MTP (6)
& 1.70$\times$ & 3.10
& 1.77$\times$ & 3.15
& 1.66$\times$ & 2.96
& 1.55$\times$ & 2.84
& 1.53$\times$ & 2.82
& 1.56$\times$ & 2.79
& 1.39$\times$ & 2.58
& 1.60$\times$ & 2.89 \\
Vanilla DFlash (6)
& 1.87$\times$ & 2.43
& 1.66$\times$ & 2.13
& 1.49$\times$ & 1.91
& 1.55$\times$ & 1.98
& 1.56$\times$ & 1.98
& 1.40$\times$ & 1.80
& 1.43$\times$ & 1.82
& 1.57$\times$ & 2.01 \\
HyperDFlash (6)
& \best{3.17$\times$} & \best{4.34}
& \best{2.78$\times$} & \best{3.67}
& \best{2.41$\times$} & \best{3.15}
& \best{2.51$\times$} & \best{3.33}
& \best{2.56$\times$} & \best{3.35}
& \best{2.18$\times$} & \best{2.88}
& \best{2.14$\times$} & \best{2.79}
& \best{2.53$\times$} & \best{3.36} \\
\midrule
Temperature = 1
& Speedup & $\tau$
& Speedup & $\tau$
& Speedup & $\tau$
& Speedup & $\tau$
& Speedup & $\tau$
& Speedup & $\tau$
& Speedup & $\tau$
& Speedup & $\tau$ \\
\midrule
MTP (3)
& 2.05$\times$ & 2.78
& 2.12$\times$ & 2.75
& 2.01$\times$ & 2.59
& 1.90$\times$ & 2.53
& 1.91$\times$ & 2.48
& 1.89$\times$ & 2.44
& 1.81$\times$ & 2.33
& 1.96$\times$ & 2.56 \\
MTP (6)
& 1.62$\times$ & 2.95
& 1.62$\times$ & 2.90
& 1.48$\times$ & 2.66
& 1.42$\times$ & 2.61
& 1.42$\times$ & 2.54
& 1.40$\times$ & 2.51
& 1.28$\times$ & 2.39
& 1.46$\times$ & 2.65 \\
Vanilla DFlash (6)
& 1.80$\times$ & 2.33
& 1.59$\times$ & 2.04
& 1.43$\times$ & 1.83
& 1.47$\times$ & 1.89
& 1.46$\times$ & 1.87
& 1.34$\times$ & 1.71
& 1.36$\times$ & 1.73
& 1.49$\times$ & 1.91 \\
HyperDFlash (6)
& \best{2.85$\times$} & \best{3.88}
& \best{2.45$\times$} & \best{3.23}
& \best{2.15$\times$} & \best{2.81}
& \best{2.22$\times$} & \best{2.94}
& \best{2.19$\times$} & \best{2.87}
& \best{1.95$\times$} & \best{2.55}
& \best{1.93$\times$} & \best{2.52}
& \best{2.25$\times$} & \best{2.97} \\
\bottomrule
\end{tabular}
}
\end{table*}

\section{Experiments}
\label{sec:results}

\subsection{Datasets and Hyperparameters}
\label{sec:datasets}
We train the drafter in two stages.
Stage 1 trains a general-purpose drafter on about 300K examples from public instruction, dialogue, and code data, with the public portion mainly based on EagleChat~\cite{zhaode2026eaglechat}.

Stage 2 continues training from the Stage 1 checkpoint on about 150K task-oriented examples, including code and instruction data such as Evol-CodeAlpaca~\cite{theblackcat2024evolcodealpaca}.
This second stage adapts the drafter to the math, code, and chat-style generation scenarios used in our evaluation.
Both stages are trained for 5 epochs on 8 NVIDIA H20 GPUs with a per-GPU batch size of 4.
The learning rate is $8{\times}10^{-4}$ for Stage 1 and $1{\times}10^{-4}$ for Stage 2.
The final checkpoint used for evaluation is selected after the second-stage adaptation.

\subsection{Benchmarks}
\label{sec:benchmarks}
We evaluate a single DeepSeek-V4-Flash target~\cite{deepseekai2026v4flash} on a benchmark suite covering math, code, and chat generation.
The math group contains GSM8K~\cite{gsm8k}, MATH-500~\cite{math500}, and AIME25; the code group contains HumanEval~\cite{humaneval}, MBPP~\cite{mbpp}, and LiveCodeBench (LCB)~\cite{jain2025livecodebench}; and the chat group contains MT-Bench~\cite{mtbench}.
This grouping separates short-form arithmetic, competition-style math, program synthesis, benchmark-style code completion, live coding tasks, and open-ended conversational generation.

All methods are evaluated under the same decoding and evaluation configuration, and all throughput and speedup measurements are conducted on the vLLM~\cite{kwon2023efficient} inference stack.
We report results for both Non-think and Think-high modes, with temperature set to 0 and 1.

\subsection{Decoding Configurations}
\label{sec:decoding-configs}
We evaluate four drafting settings under the same target model and verification pipeline, varying only the drafting mechanism and the number of drafted steps per verification round.
MTP (3) is the native MTP deployment point with three drafted steps, while MTP (6) keeps the native MTP mechanism but matches the six-step draft budget used by the DFlash variants.
Vanilla DFlash (6) is the generic six-step DFlash adaptation.
HyperDFlash (6) enables the DeepSeek-specific conditioning source, inherited path reducer, and KL regularization described in Section~\ref{sec:method}.
Parenthesized values indicate the number of drafted steps per verification round.

\subsection{Metrics}
\label{sec:metrics}
We report decoding speedup over target-only autoregressive decoding and the mean accepted length $\tau$.
Speedup measures the relative generation throughput of each speculative decoding configuration against the same target-only baseline.
The accepted length $\tau$ is the average number of draft tokens accepted per verification round and is the direct acceptance-quality signal used throughout the report.

\subsection{Main Results}
\label{sec:main-results}
Tables~\ref{tab:main-results-nonthink} and~\ref{tab:main-results-thinkhigh} report the main decoding results across benchmark groups, reasoning modes, and sampling temperatures.
HyperDFlash consistently improves over the native MTP deployment setting.
In non-thinking mode at temperature 0, it increases the average accepted length from 2.93 to 3.69 over MTP (3), with the average decoding speedup improving from 2.25$\times$ to 2.80$\times$.

The gain remains clear under the same six-step draft budget.
Compared with MTP (6), HyperDFlash improves the average accepted length from 3.08 to 3.69 in non-thinking mode at temperature 0, while also increasing the average speedup from 1.76$\times$ to 2.80$\times$.
This shows that the improvement does not simply come from drafting more tokens.

HyperDFlash also substantially outperforms Vanilla DFlash (6), whose average accepted length is only 2.14 under the same non-thinking temperature-0 setting.
This gap indicates that a direct DFlash adaptation is insufficient for DeepSeek-V4, and that \hc{}-aware conditioning and path reduction are important for effective block-parallel speculative decoding.

\section{Related Work}
\label{sec:related}
Speculative decoding accelerates autoregressive generation by using a cheaper proposal mechanism to draft future tokens and then verifying those tokens with the target model in parallel~\cite{leviathan2023,chen2023}.
Subsequent work extends this draft-and-verify template with token-tree verification, retrieval drafting, lookahead candidates, or self-speculative early exits~\cite{miao2024specinfer,he2024rest,fu2024lookahead,zhang2024draftverify,elhoushi2024layerskip}.
These methods reduce target-model decoding steps, but they differ in how draft tokens are proposed and what serving or training infrastructure is required.

Learned drafters train auxiliary prediction modules to improve acceptance.
Blockwise parallel decoding predicts multiple tokens per pass~\cite{stern2018}, Medusa and Hydra attach multi-token draft heads~\cite{cai2024medusa,ankner2024hydra}, and EAGLE trains feature-level speculative drafters with dynamic draft trees~\cite{li2024eagle,li2024eagle2,li2025eagle3}.
DFlash is closest to our setting because it drafts an entire block from target hidden-state context~\cite{chen2026dflash}.
HyperDFlash keeps this block-parallel interface but adapts it to DeepSeek-V4's \hc{} residual stream~\cite{zhu2024hyper} and native multi-token prediction setting~\cite{gloeckle2024,xu2026deepseek}.
DeepSeek-V4 implements this hyper-connection design as mHC~\cite{xu2026deepseek}.
Our auxiliary KL term follows standard knowledge distillation~\cite{hinton2015}, but is applied conservatively because later draft positions condition on a different information set from the teacher.

\section{Conclusion}
\label{sec:conclusion}
This work presents HyperDFlash, a hyper-connection-aligned block-parallel speculative decoding framework tailored for DeepSeek-V4. We systematically resolve the structural mismatch between conventional block drafting and \hc{} residual streams via three compact, coordinated designs: pre-collapse residual conditioning for strict target alignment, inherited gated reduction for ultra-efficient path aggregation, and early-stage KL distillation for high-quality draft refinement. Evaluations across math reasoning, code synthesis, and conversational tasks verify that HyperDFlash surpasses both native MTP and vanilla DFlash in accepted token length and decoding speed. Our architecture-aware drafting paradigm offers an effective, lightweight solution for accelerating modern hyper-connection-based large language models.

\section*{Limitations}
Although we report decoding speedup, a more complete serving study should further separate drafter cost, verification cost, batching effects, and end-to-end latency under production traffic; we leave a full production-level latency comparison against MTP to future work.
The KL objective is motivated by the position-wise analysis in Section~\ref{sec:kl} but is not ablated in isolation; moreover, the current teacher mean-pools the \hc{} paths instead of applying the gated \hchead{} collapse, a bias we mitigate by keeping its weight small.
The inherited reducer assumes the \prehc{} source and matching target and draft hidden widths, and falls back to a generic projection otherwise.
Finally, all results are obtained on a single target model and include an internal workload that is not publicly reproducible; validating the recipe on other \hc{}-based targets is left to future work.

\bibliographystyle{plainnat}
\bibliography{sample}

@article{chen2026dflash,
  title   = {{DFlash}: Block Diffusion for Flash Speculative Decoding},
  author  = {Chen, Jian and Liang, Yesheng and Liu, Zhijian},
  journal = {arXiv preprint arXiv:2602.06036},
  year    = {2026}
}

@inproceedings{leviathan2023,
  title     = {Fast Inference from Transformers via Speculative Decoding},
  author    = {Leviathan, Yaniv and Kalman, Matan and Matias, Yossi},
  booktitle = {Proceedings of the 40th International Conference on Machine Learning (ICML)},
  year      = {2023}
}

@article{chen2023,
  title   = {Accelerating Large Language Model Decoding with Speculative Sampling},
  author  = {Chen, Charlie and Borgeaud, Sebastian and Irving, Geoffrey and Lespiau, Jean-Baptiste and Sifre, Laurent and Jumper, John},
  journal = {arXiv preprint arXiv:2302.01318},
  year    = {2023}
}

@inproceedings{cai2024medusa,
  title     = {Medusa: Simple LLM Inference Acceleration Framework with Multiple Decoding Heads},
  author    = {Cai, Tianle and Li, Yuhong and Geng, Zhengyang and Peng, Hongwu and Lee, Jason D. and Chen, Deming and Dao, Tri},
  booktitle = {Proceedings of the 41st International Conference on Machine Learning (ICML)},
  year      = {2024}
}

@inproceedings{li2024eagle,
  title     = {EAGLE: Speculative Sampling Requires Rethinking Feature Uncertainty},
  author    = {Li, Yuhui and Wei, Fangyun and Zhang, Chao and Zhang, Hongyang},
  booktitle = {Proceedings of the 41st International Conference on Machine Learning (ICML)},
  year      = {2024}
}

@inproceedings{li2024eagle2,
  title     = {EAGLE-2: Faster Inference of Language Models with Dynamic Draft Trees},
  author    = {Li, Yuhui and Wei, Fangyun and Zhang, Chao and Zhang, Hongyang},
  booktitle = {Proceedings of the 2024 Conference on Empirical Methods in Natural Language Processing (EMNLP)},
  year      = {2024}
}

@article{li2025eagle3,
  title   = {EAGLE-3: Scaling up Inference Acceleration of Large Language Models via Training-Time Test},
  author  = {Li, Yuhui and Wei, Fangyun and Zhang, Chao and Zhang, Hongyang},
  journal = {arXiv preprint arXiv:2503.01840},
  year    = {2025}
}

@inproceedings{stern2018,
  title     = {Blockwise Parallel Decoding for Deep Autoregressive Models},
  author    = {Stern, Mitchell and Shazeer, Noam and Uszkoreit, Jakob},
  booktitle = {Advances in Neural Information Processing Systems (NeurIPS)},
  year      = {2018}
}

@inproceedings{gloeckle2024,
  title     = {Better \& Faster Large Language Models via Multi-token Prediction},
  author    = {Gloeckle, Fabian and Idrissi, Badr Youbi and Rozi{\`e}re, Baptiste and Lopez-Paz, David and Synnaeve, Gabriel},
  booktitle = {Proceedings of the 41st International Conference on Machine Learning (ICML)},
  year      = {2024}
}

@article{zhu2024hyper,
  title   = {Hyper-Connections},
  author  = {Zhu, Defa and Huang, Hongzhi and Huang, Zihao and Zeng, Yutao and Mao, Yunyao and Wu, Banggu and Min, Qiyang and Zhou, Xun},
  journal = {arXiv preprint arXiv:2409.19606},
  year    = {2024}
}

@article{hinton2015,
  title   = {Distilling the Knowledge in a Neural Network},
  author  = {Hinton, Geoffrey and Vinyals, Oriol and Dean, Jeff},
  journal = {arXiv preprint arXiv:1503.02531},
  year    = {2015}
}

@inproceedings{miao2024specinfer,
  title     = {{SpecInfer}: Accelerating Generative Large Language Model Serving with Tree-based Speculative Inference and Verification},
  author    = {Miao, Xupeng and Oliaro, Gabriele and Zhang, Zhihao and Cheng, Xinhao and Wang, Zeyu and Zhang, Zhengxin and Wong, Rae Ying Yee and Zhu, Alan and Yang, Lijie and Shi, Xiaoxiang and Shi, Chunan and Chen, Zhuoming and Arfeen, Daiyaan and Abhyankar, Reyna and Jia, Zhihao},
  booktitle = {Proceedings of the 29th ACM International Conference on Architectural Support for Programming Languages and Operating Systems (ASPLOS)},
  year      = {2024}
}

@inproceedings{he2024rest,
  title     = {{REST}: Retrieval-Based Speculative Decoding},
  author    = {He, Zhenyu and Zhong, Zexuan and Cai, Tianle and Lee, Jason D. and He, Di},
  booktitle = {Proceedings of the 2024 Conference of the North American Chapter of the Association for Computational Linguistics (NAACL)},
  year      = {2024}
}

@inproceedings{fu2024lookahead,
  title     = {Break the Sequential Dependency of {LLM} Inference Using Lookahead Decoding},
  author    = {Fu, Yichao and Bailis, Peter and Stoica, Ion and Zhang, Hao},
  booktitle = {Proceedings of the 41st International Conference on Machine Learning (ICML)},
  year      = {2024}
}

@inproceedings{zhang2024draftverify,
  title     = {Draft \& Verify: Lossless Large Language Model Acceleration via Self-Speculative Decoding},
  author    = {Zhang, Jun and Wang, Jue and Li, Huan and Shou, Lidan and Chen, Ke and Chen, Gang and Mehrotra, Sharad},
  booktitle = {Proceedings of the 62nd Annual Meeting of the Association for Computational Linguistics (ACL)},
  year      = {2024}
}

@inproceedings{elhoushi2024layerskip,
  title     = {{LayerSkip}: Enabling Early Exit Inference and Self-Speculative Decoding},
  author    = {Elhoushi, Mostafa and Shrivastava, Akshat and Liskovich, Diana and Hosmer, Basil and Wasti, Bram and Lai, Liangzhen and Mahmoud, Anas and Acun, Bilge and Agarwal, Saurabh and Roman, Ahmed and Aly, Ahmed A. and Chen, Beidi and Wu, Carole-Jean},
  booktitle = {Proceedings of the 62nd Annual Meeting of the Association for Computational Linguistics (ACL)},
  year      = {2024}
}

@article{ankner2024hydra,
  title   = {{Hydra}: Sequentially-Dependent Draft Heads for Medusa Decoding},
  author  = {Ankner, Zachary and Parthasarathy, Rishab and Nrusimha, Aniruddha and Rinard, Christopher and Ragan-Kelley, Jonathan and Brandon, William},
  journal = {arXiv preprint arXiv:2402.05109},
  year    = {2024}
}

@article{gsm8k,
  title={Training verifiers to solve math word problems},
  author={Cobbe, Karl and Kosaraju, Vineet and Bavarian, Mohammad and Chen, Mark and Jun, Heewoo and Kaiser, Lukasz and Plappert, Matthias and Tworek, Jerry and Hilton, Jacob and Nakano, Reiichiro and others},
  journal={arXiv preprint arXiv:2110.14168},
  year={2021}
}

@article{math500,
  title={Measuring mathematical problem solving with the math dataset},
  author={Hendrycks, Dan and Burns, Collin and Kadavath, Saurav and Arora, Akul and Basart, Steven and Tang, Eric and Song, Dawn and Steinhardt, Jacob},
  journal={arXiv preprint arXiv:2103.03874},
  year={2021}
}

@article{humaneval,
  title={Evaluating large language models trained on code},
  author={Chen, Mark and Tworek, Jerry and Jun, Heewoo and Yuan, Qiming and Pinto, Henrique Ponde De Oliveira and Kaplan, Jared and Edwards, Harri and Burda, Yuri and Joseph, Nicholas and Brockman, Greg and others},
  journal={arXiv preprint arXiv:2107.03374},
  year={2021}
}

@article{mbpp,
  title={Program synthesis with large language models},
  author={Austin, Jacob and Odena, Augustus and Nye, Maxwell and Bosma, Maarten and Michalewski, Henryk and Dohan, David and Jiang, Ellen and Cai, Carrie and Terry, Michael and Le, Quoc and others},
  journal={arXiv preprint arXiv:2108.07732},
  year={2021}
}

@inproceedings{jain2025livecodebench,
  title={Livecodebench: Holistic and contamination free evaluation of large language models for code},
  author={Jain, Naman and Gu, Alex and Li, Wen-Ding and Yan, Fanjia and Zhang, Tianjun and Wang, Sida and Solar-Lezama, Armando and Sen, Koushik and Stoica, Ion},
  booktitle={International Conference on Learning Representations},
  volume={2025},
  pages={58791--58831},
  year={2025}
}

@article{mtbench,
  title={Judging llm-as-a-judge with mt-bench and chatbot arena},
  author={Zheng, Lianmin and Chiang, Wei-Lin and Sheng, Ying and Zhuang, Siyuan and Wu, Zhanghao and Zhuang, Yonghao and Lin, Zi and Li, Zhuohan and Li, Dacheng and Xing, Eric and others},
  journal={Advances in neural information processing systems},
  volume={36},
  pages={46595--46623},
  year={2023}
}

@misc{deepseekai2026v4flash,
  title        = {{DeepSeek-V4-Flash}},
  author       = {{DeepSeek-AI}},
  howpublished = {\url{https://huggingface.co/deepseek-ai/DeepSeek-V4-Flash}},
  year         = {2026},
  note         = {Hugging Face model card}
}

@misc{zhaode2026eaglechat,
  title        = {{EagleChat}},
  author       = {Zhao, De},
  howpublished = {\url{https://huggingface.co/datasets/zhaode/EagleChat}},
  year         = {2026},
  note         = {Hugging Face dataset card}
}

@misc{theblackcat2024evolcodealpaca,
  title        = {{Evol-CodeAlpaca-v1}},
  author       = {{theblackcat102}},
  howpublished = {\url{https://huggingface.co/datasets/theblackcat102/evol-codealpaca-v1}},
  year         = {2024},
  note         = {Hugging Face dataset card}
}

@article{xu2026deepseek,
  title={DeepSeek-V4: Towards Highly Efficient Million-Token Context Intelligence},
  author={Xu, Anyi and Lin, Bangcai and Xue, Bing and Wang, Bingxuan and Xu, Bingzheng and Wu, Bochao and Zhang, Bowei and Lin, Chaofan and Dong, Chen and Ling, Chenchen and others},
  journal={arXiv preprint arXiv:2606.19348},
  year={2026}
}

@inproceedings{kwon2023efficient,
  title={Efficient Memory Management for Large Language Model Serving with PagedAttention},
  author={Woosuk Kwon and Zhuohan Li and Siyuan Zhuang and Ying Sheng and Lianmin Zheng and Cody Hao Yu and Joseph E. Gonzalez and Hao Zhang and Ion Stoica},
  booktitle={Proceedings of the ACM SIGOPS 29th Symposium on Operating Systems Principles},
  year={2023}
}

\end{document}